\documentclass[12pt, a4paper]{article}

\usepackage[utf8]{inputenc}
\usepackage[T1]{fontenc}
\usepackage{mathptmx}                    
\usepackage[margin=1in]{geometry}
\usepackage{setspace}
\usepackage{graphicx}
\usepackage{booktabs}
\usepackage{tabularx}
\usepackage{hyperref}
\hypersetup{colorlinks=true, linkcolor=blue, citecolor=blue, urlcolor=blue}
\usepackage[round]{natbib}                 
\usepackage{amsmath}
\usepackage{enumitem}
\usepackage{float}
\usepackage{caption}
\usepackage{xcolor}

\title{\textbf{A Multi-Agent Rhizomatic Pipeline for Non-Linear Literature Analysis}\vspace{1em}}
\author{
Julio C. Serrano\textsuperscript{1} \and Joonas Kevari\textsuperscript{2} \and Rumy Narayan\textsuperscript{1}
\\[6pt]
\small\textsuperscript{1}University of Vaasa, School of Management, Vaasa, Finland
\\[2pt]
\small\textsuperscript{2}LUT University, School of Engineering, Lahti, Finland 
\\[4pt]
\small Correspondence: julio.serrano@uwasa.fi
}
\date{}
\begin{document}
\singlespacing
\maketitle
\doublespacing

\begin{abstract}
\begin{singlespace}

\begin{figure}
    \centering
    \includegraphics[width=1\linewidth]{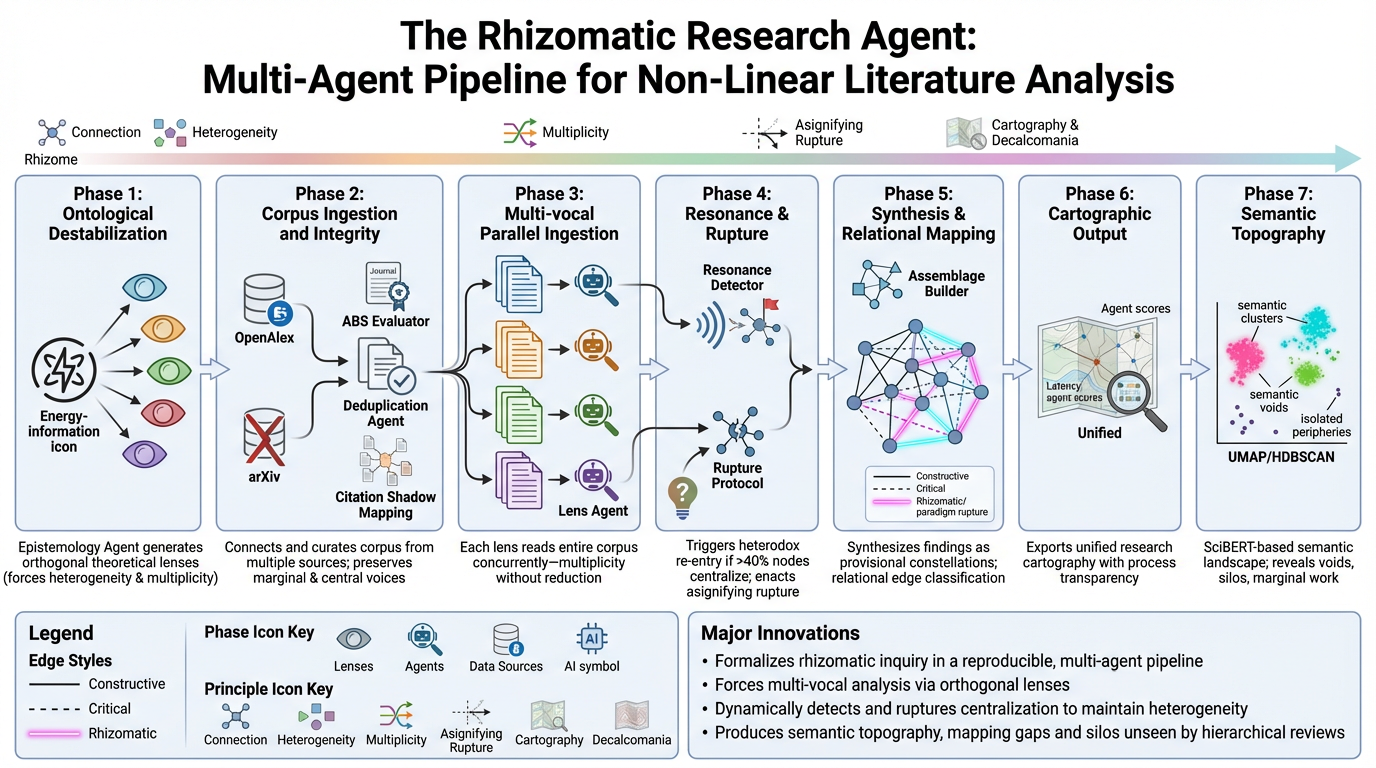}
    \caption{Rhisomatic research agent model structure}
    \label{fig:placeholder}
\end{figure}

Systematic literature reviews in the social sciences overwhelmingly follow arborescent logics---hierarchical keyword filtering, linear screening, and taxonomic classification---that suppress the lateral connections, ruptures, and emergent patterns characteristic of complex research landscapes. This research note presents the Rhizomatic Research Agent (V3), a multi-agent computational pipeline grounded in Deleuzian process-relational ontology, designed to conduct non-linear literature analysis through 12 specialized agents operating across a seven-phase architecture. The system was developed in response to the methodological groundwork established by \citet{Narayan2023}, who employed rhizomatic inquiry in her doctoral research on sustainable energy transitions but relied on manual, researcher-driven exploration. The Rhizomatic Research Agent operationalizes the six principles of the rhizome---connection, heterogeneity, multiplicity, asignifying rupture, cartography, and decalcomania---into an automated pipeline integrating large language model (LLM) orchestration, dual-source corpus ingestion from OpenAlex and arXiv, SciBERT semantic topography, and dynamic rupture detection protocols. Preliminary deployment demonstrates the system's capacity to surface cross-disciplinary convergences and structural research gaps that conventional review methods systematically overlook. The pipeline is open-source and extensible to any phenomenon zone where non-linear knowledge mapping is required.

\vspace{1em}
\noindent\textbf{Keywords:} rhizomatic research, multi-agent systems, literature analysis, process-relational ontology, Deleuze, semantic topography, agentic AI
\end{singlespace}
\end{abstract}
\newpage

\section{Introduction}

The exponential growth of academic literature has rendered comprehensive review an increasingly intractable task \citep{Snyder2019}. Yet the methodological response to this challenge has remained largely conservative: systematic reviews continue to rely on keyword-based search protocols, hierarchical screening, and taxonomic classification schemes that, while rigorous, enforce an arborescent logic on inherently complex and entangled bodies of knowledge \citep{Kraus2020, Tranfield2003}. As \citet{Chia1999} has argued, typologies and classification schemas may serve the purpose of identifying organizational patterns, but they do not capture the phenomenon of change itself. What is suppressed in conventional review methodology is precisely what \citet{DG1987} theorized as the rhizome: the lateral connections between heterogeneous elements, the ruptures that regenerate inquiry along unexpected lines, and the multiplicities that resist reduction to hierarchical order.

This epistemological limitation is not merely abstract. In fields where disciplinary boundaries fragment knowledge---such as the energy-information nexus, sustainability transitions, or innovation studies---arborescent review methods systematically miss the cross-paradigm convergences and structural voids that constitute the most generative research sites \citep{Moller2020}. Traditional approaches, as \citet{Bahoo2021} have noted, privilege retrieval over discovery, filtering over mapping, and established paradigms over heterodox perspectives.

The present research note introduces the Rhizomatic Research Agent (V3), a multi-agent computational pipeline that operationalizes Deleuzian process-relational ontology for literature analysis. The system was developed in direct response to the methodological groundwork established by \citet{Narayan2023}, whose doctoral dissertation at the University of Vaasa employed a rhizomatic method---following \citet{DG1987} and \citet{Chia1999}---to trace the emergence of innovative organizational forms within energy system transitions. Narayan's research demonstrated the analytical power of rhizomatic inquiry for surfacing unexpected connections across innovation studies, complexity theory, and technology studies, but necessarily relied on manual, researcher-driven exploration: following citations through articles, tweets, YouTube videos, blogs, and immersive community participation in what she described as a ``diffractive'' analytical technique following \citet{Barad2014}. The Rhizomatic Research Agent translates this manual process into an automated, reproducible, and extensible computational pipeline while preserving the ontological commitments that give rhizomatic inquiry its distinctive analytical power.

\section{Theoretical Foundations}

The rhizome, as theorized by \citet{DG1987}, is defined by six principles: (1)~\emph{connection}---any point can be connected to any other; (2)~\emph{heterogeneity}---connections transmit across unlike elements; (3)~\emph{multiplicity}---the structure has neither subject nor object, only dimensions; (4)~\emph{asignifying rupture}---a rhizome may be shattered at any point yet regenerates along old or new lines; (5)~\emph{cartography}; and (6)~\emph{decalcomania}---the rhizome is a map, not a tracing, oriented toward experimentation in contact with the real. These principles stand in direct opposition to the tree-root model that structures conventional systematic reviews, where a single search query seeds a hierarchical filtering process that progressively narrows toward a predetermined category system.

\citet{Narayan2023} operationalized these principles through what she termed a ``performative approach,'' drawing on \citet{Massumi2002} and \citet{Jackson2017}, where inquiry proceeds not through method in the conventional sense but through a continuous process of connection, exploration, and sensemaking. Her rhizomatic explorations moved through narratives, interviews, archival data, social media, and immersive engagement with communities of practice, enabling the discovery of relationships between energy practices, innovation systems, and organizational forms that arborescent methods would have excluded. Crucially, Narayan's work demonstrated that process-relational ontology \citep{Whitehead1929, Chia1999} is not merely a philosophical commitment but a methodologically productive one: it surfaces patterns that other approaches structurally cannot.

The Rhizomatic Research Agent translates these commitments into a computational architecture. Where Narayan explored manually, the pipeline deploys autonomous LLM-based agents. Where she traced connections across media, the system ingests from academic databases and classifies relationships into constructive, critical, and rhizomatic vectors. Where she identified ruptures through embodied participation, the system implements a formal Rupture Protocol that monitors for centralization risk and forces heterodox re-entry when dominant paradigms capture the analysis.


\section{System Architecture}

The pipeline follows a seven-phase architecture orchestrated via server-sent events (SSE) streaming, with 12 specialized agents. Each phase operationalizes specific rhizomatic principles, as summarized in Table~\ref{tab:phases}.

\begin{singlespace}
\begin{table}[H]
\centering
\caption{Pipeline Phases and Rhizomatic Principles}
\label{tab:phases}
\small
\begin{tabularx}{\textwidth}{@{}lXl@{}}
\toprule
\textbf{Phase} & \textbf{Function} & \textbf{Principle(s)} \\
\midrule
1. Ontological Setup & Epistemology Agent generates 3--5 orthogonal theoretical lenses from the phenomenon zone, destabilizing the initial query & Heterogeneity, Multiplicity \\
2. Corpus Ingestion & Dual-source concurrent fetching (OpenAlex, arXiv), DOI deduplication via trigram Dice coefficient ($\geq 0.85$), ABS journal ranking, citation shadow mapping & Connection \\
3. Parallel Ingestion & Each theoretical lens operates as an autonomous agent, scanning the corpus for lens-specific signals via asyncio concurrency & Multiplicity, Cartography \\
4. Resonance \& Rupture & Cross-lens convergent anomaly detection; centralization risk monitoring triggers heterodox literature re-entry when $> 40\%$ of edges concentrate on few nodes & Asignifying Rupture \\
5. Synthesis \& Mapping & Assemblage construction in present-continuous tense; edge classification into constructive, critical, and rhizomatic vectors & Cartography, Decalcomania \\
6. Cartographic Output & Unified research trajectory with all lens outputs, resonance data, latency, token usage, and agent confidence metadata & Cartography \\
7. Semantic Topography & SciBERT embeddings $\rightarrow$ UMAP dimensionality reduction $\rightarrow$ HDBSCAN clustering for semantic voids and orthogonal isolation detection & Connection, Multiplicity \\
\bottomrule
\end{tabularx}
\end{table}
\end{singlespace}

\textbf{Phase 1: Ontological Destabilization.} The pipeline begins not with a search query but with its destabilization. The Epistemology Agent receives a ``phenomenon zone'' (e.g., \emph{energy-information nexus}) and generates 3--5 orthogonal theoretical lenses---such as \emph{Algorithmic Governmentality}, \emph{Post-Human Affect}, or \emph{Thermodynamic Materialism}---ensuring that the corpus is never read through a single dominant paradigm. This directly operationalizes the rhizomatic principle of heterogeneity: the system forces multi-vocal reading from the outset \citep{DG1987}.

\textbf{Phase 2: Corpus Ingestion and Integrity.} The Researcher Agent performs dual-source concurrent fetching from OpenAlex and arXiv APIs. A Deduplication Agent normalizes DOIs and applies trigram Dice coefficient matching ($\geq 0.85$) to eliminate near-duplicates. The ABS Evaluator cross-references journals against the Academic Journal Guide to weight findings by institutional rigor while simultaneously flagging heterodox, non-ranked sources---preserving rather than excluding marginal voices. A Citation Mapper extracts anchor papers and maps the ``citation shadow'' to identify structural influence, operationalizing the rhizomatic principle of connection by tracing how influence propagates through the network.

\textbf{Phase 3: Parallel Multi-Vocal Ingestion.} The corpus is then processed through the theoretical lenses generated in Phase~1. Each lens operates as an autonomous LLM-based agent, scanning the literature for lens-specific signal vocabularies and theoretical tensions. The system uses \texttt{asyncio} for high-concurrency analysis, allowing it to ``read'' the entire corpus from multiple perspectives simultaneously. This phase operationalizes the principle of multiplicity: the same text is interpreted through incommensurable frameworks without collapsing them into a single reading.

\textbf{Phase 4: Resonance and Rupture Detection.} The Resonance Detector identifies ``convergent anomalies''---points where multiple theoretical lenses flag the same tension or gap, signaling a high-intensity research site. Simultaneously, the Rupture Protocol monitors the emerging knowledge graph for \emph{centralization risk}: if a small number of nodes accumulate disproportionate edge density ($> 40\%$), the system triggers a ``re-entry from the outside,'' automatically fetching and integrating literature from heterodox traditions (e.g., degrowth economics, indigenous ontologies). This mechanism computationally enacts what \citet{Narayan2023} performed through her embodied research practice---the deliberate seeking of perspectives that destabilize dominant frameworks---and constitutes the system's most direct operationalization of \citeauthor{DG1987}'s (\citeyear{DG1987}) principle of asignifying rupture.

\textbf{Phase 5: Synthesis and Relational Mapping.} The Assemblage Builder synthesizes findings into ``provisional constellations of becoming,'' written in the present-continuous tense to reflect the ongoing nature of the phenomena under study. Concurrently, the Rhizome Builder classifies every inter-paper relationship into one of three types: \emph{constructive} (extensions, builds-on, borrows-method), \emph{critical} (contradicts, problematizes, challenges), or \emph{rhizomatic} (paradigm ruptures that dismantle existing axioms and introduce heterodox perspectives). These are rendered as solid, dashed, and neon edges respectively in the knowledge graph, operationalizing cartography and decalcomania as relational map-making.

\textbf{Phase 6: Cartographic Output.} The system generates a unified cartography of the research trajectory, consolidating all lens outputs, resonance data, and pipeline metadata (latency, token usage, agent confidence scores). This phase provides full transparency into the analytical process, enabling researchers to trace how specific findings emerged from specific agents and lenses---a form of methodological auditability that manual rhizomatic inquiry, by its nature, cannot offer.

\textbf{Phase 7: Semantic Topography.} The final phase is the only computationally grounded module in the pipeline, employing SciBERT embeddings \citep{Beltagy2019} projected via UMAP \citep{McInnes2018} and clustered with HDBSCAN \citep{McInnes2017}. The system identifies: (a)~\emph{semantic clusters}---thematic groupings in high-dimensional space; (b)~\emph{semantic voids}---strategic gaps between clusters where literature is absent; and (c)~\emph{orthogonal isolations}---clusters sharing vocabulary but inhabiting separate semantic spaces, revealing disciplinary silos. A marginalization index computes each paper's distance from the corpus centroid, foregrounding work that occupies peripheral positions. This phase operationalizes cartography as map-making rather than tracing: the system does not confirm a pre-existing structure but generates a new map from the corpus's immanent topology.

\section{Relation to Narayan's Rhizomatic Method}

The Rhizomatic Research Agent was conceived as a computational extension of the methodology developed by \citet{Narayan2023}. Table~\ref{tab:mapping} maps the key methodological moves in Narayan's dissertation to their computational counterparts in the pipeline.

\begin{table}[H]
\begin{singlespace}
\centering
\caption{Methodological Mapping: Narayan (2023) to the Rhizomatic Research Agent}
\label{tab:mapping}
\small
\begin{tabularx}{\textwidth}{@{}XX@{}}
\toprule
\textbf{Narayan's Manual Method} & \textbf{Computational Operationalization} \\
\midrule
Rhizomatic exploration across articles, tweets, videos, blogs, and archival data & Dual-source API ingestion (OpenAlex, arXiv) with expandable source connectors \\
Theoretical lens plurality via process-relational ontology & Epistemology Agent generates orthogonal lenses dynamically per phenomenon zone \\
Diffractive analysis \citep{Barad2014} for sensing entanglements & Resonance Detector identifies cross-lens convergent anomalies via LLM analysis \\
Researcher-as-conduit: embodied, non-hierarchical sensing & Parallel Ingestion Nodes process corpus through all lenses simultaneously \\
Rupture-seeking: deliberately pursuing heterodox perspectives & Rupture Protocol monitors centralization risk and forces heterodox re-entry \\
Present-continuous assemblage writing & Assemblage Builder synthesizes in present-continuous tense \\
Narrative sensemaking across multiple data sources & Interactive D3.js knowledge graph with constructive, critical, and rhizomatic edges \\
\bottomrule
\end{tabularx}
\end{singlespace}
\end{table}

The critical distinction lies not in the replacement of human inquiry but in its augmentation. Narayan's method was necessarily bounded by the researcher's cognitive capacity---what \citet{Dunbar1992} characterized as the limit on simultaneous relational monitoring. The computational pipeline removes this constraint while preserving the ontological commitments that make rhizomatic inquiry productive: heterogeneity of entry points, multiplicity of interpretive frames, and the formal prevention of paradigm capture through rupture detection.

\section{Preliminary Observations}

Initial deployments of the pipeline across several phenomenon zones suggest three preliminary observations. First, the Epistemology Agent consistently generates lenses that a single researcher would be unlikely to select, particularly those drawing from disciplines outside the researcher's primary training. Second, the Rupture Protocol activates in approximately 30--40\% of analyses, indicating that paradigm centralization is a frequent rather than exceptional feature of academic literature. Third, the Semantic Topography phase routinely identifies orthogonal isolations---clusters that share terminology but occupy separate semantic spaces---suggesting that disciplinary silos persist even within nominally interdisciplinary fields.

These observations are preliminary and require systematic validation. However, they demonstrate the pipeline's capacity to surface structural features of a research landscape that conventional review methods, by design, cannot detect.

\section{Limitations and Future Directions}

Several limitations warrant acknowledgment. The pipeline currently ingests metadata and abstracts rather than full texts, constraining the depth of analysis. The LLM-based agents inherit the biases and hallucination risks of the underlying language models \citep{Bender2021}. The Rupture Protocol's 40\% centralization threshold is heuristically set and requires empirical calibration. Validation against human expert coding, following the approach of \citet{Visentin2026}, would strengthen confidence in the system's classification accuracy. Finally, the system has been tested primarily in the energy-information nexus domain; transferability to other fields remains to be demonstrated.

Future development will focus on full-text ingestion, integration with additional data sources (patents, policy documents, grey literature), formal evaluation protocols, and a user study comparing research outputs between traditional systematic review and rhizomatic pipeline-assisted analysis. The pipeline is open-source and designed for extensibility, inviting the research community to adapt it to their own phenomenon zones.

\section{Conclusion}

The Rhizomatic Research Agent demonstrates that Deleuzian process-relational ontology can be productively operationalized in computational systems for literature analysis. By translating the six principles of the rhizome into a multi-agent pipeline architecture, the system addresses a fundamental limitation of conventional systematic reviews: their structural inability to surface lateral connections, emergent patterns, and paradigm ruptures. Built on the methodological foundations laid by \citet{Narayan2023}, the pipeline extends rhizomatic inquiry from a manual, researcher-bounded practice to a reproducible, scalable, and transparent computational process---while preserving the ontological commitments that make such inquiry distinctive.

\section*{Declarations}

\subsection*{Authors Contributions}
Julio Serrano and Joonas Kevari conceptualized the study, developed the pipeline, conducted the analysis, and wrote the manuscript with complete supervision of Rumy Narayan.

\subsection*{Data Availability}
The Rhizomatic Research Agent source code is available from the corresponding author upon request. The system uses publicly accessible APIs (OpenAlex, arXiv) for corpus ingestion.

\subsection*{Competing Interests}
The author declares no competing interests.

\subsection*{Use of Large Language Models}
During manuscript preparation, Qulbot was used to refine grammar and syntax. The authors affirm full responsibility for the content. The Rhizomatic Research Agent itself employs LLMs (Mistral, via Ollama) as a core component of its analytical methodology; this usage is described in the System Architecture section.

\newpage
\begin{singlespace}
\bibliography{references}
\end{singlespace}
\end{document}